\documentclass[letterpaper, 10 pt, conference]{ieeeconf}
\usepackage[utf8]{inputenc}
\usepackage[T1]{fontenc}

\IEEEoverridecommandlockouts
\usepackage{tabularx} 
\usepackage{amsmath}
\usepackage{amssymb}
\usepackage{graphicx}
\usepackage{algorithm}
\usepackage{algorithmic}
\usepackage{booktabs}
\usepackage{multirow}
\usepackage{xcolor}
\usepackage{hyperref}
\usepackage{subcaption}
\usepackage{tikz}
\usepackage{tcolorbox, listings}
\usetikzlibrary{automata, positioning, arrows.meta}

\DeclareMathOperator*{\argmax}{arg\,max}
\DeclareMathOperator*{\argmin}{arg\,min}

\newcommand{\ltlnext}{\bigcirc}
\newcommand{\ltlevent}{\Diamond}

\newcommand{\ltluntil}{\;\mathcal{U}\;}

\title{\LARGE \bf

ViTL: Temporal Logic-Guided Zero-Shot Natural Language Navigation via Vision-Language Models
}

\author{Kaier Liang$^*$, Hengde Dai$^*$ and Cristian-Ioan Vasile%
\thanks{$^*$ The authors contributed equally.}%
\thanks{Kaier Liang, Hengde Dai and Cristian-Ioan Vasile are with Lehigh University, PA, USA: {\tt\small \{kal221, hed424, cvasile\}@lehigh.edu}}
\thanks{This project has been partially supported by the NSF IIS 2442644 award.}
}

\begin{document}

\maketitle
\thispagestyle{empty}
\pagestyle{empty}

\begin{abstract}
Enabling robots to follow natural language commands to complete zero-shot long-horizon tasks
remains challenging. It requires extracting implicit temporal and logical constraints from natural language commands and executing multiple sub-tasks accordingly. Recent zero-shot object navigation methods use vision-language models~(VLMs) to guide frontier-based exploration in unknown environments, but they are limited to single-target tasks. Real-world commands such as \textit{``Clean either the chair or the couch, then turn on the tv.''} require navigating to multiple targets in a temporally constrained order, which no existing zero-shot system can handle. We present ViTL, a framework that addresses this gap at two levels. At the task level, we use a large language model~(LLM) to compile natural language commands into Linear Temporal Logic~(LTL) formulas, which are then converted into Deterministic Finite Automata~(DFA) that coordinate multi-channel value maps and trigger dynamic replanning when new objects are detected. At the navigation level, we introduce directional score: rather than producing a direction-agnostic value across the entire field of view, we label frontier directions on the observation image and extract per-direction scores from the VLM. Experiments on Habitat-Matterport 3D (HM3D) show that the full framework enables zero-shot long-horizon completion of natural language navigation tasks with temporal constraints, and that directional score improves single-target navigation accuracy and efficiency over the baseline.

\end{abstract}


\section{INTRODUCTION}
\label{sec:introduction}

Object-goal navigation (ObjectNav), the task of navigating to an object specified by its semantic category in an unknown environment, is a fundamental capability for autonomous mobile robots~\cite{chaplot2020semexp, ramakrishnan2022poni}. Recent zero-shot approaches have made remarkable progress by leveraging pre-trained vision-language models~(VLMs) to guide exploration without task-specific training~\cite{gadre2022clip, yu2023l3mvn, yokoyama2024vlfm, yin2024sg}. 

However, current zero-shot ObjectNav methods are limited to single-target navigation. Real-world robot deployments demand capabilities beyond them. For instance, consider a household service robot that receives a natural language command: \textit{``Clean either the chair or the couch, then turn on the tv.''}. Such tasks require navigating to multiple targets in a temporally constrained order. To the best of the authors' knowledge, no existing zero-shot navigation system can handle this problem.

Solving this problem requires addressing two interleaved challenges. At the \textit{task level}, the system needs a high-level planner that continuously incorporates environment observations, identifies the efficient completion path, and dynamically selects the current sub-task. At the \textit{navigation level}, each sub-task demands accurate and efficient zero-shot single-target navigation executor.

We present \textbf{ViTL} (\textbf{Vi}sion-Language \textbf{T}emporal \textbf{L}ogic Navigation), a zero-shot navigation framework that addresses both levels.
 At the \textit{task level}, we leverage prior work on natural language to temporal logic translation~\cite{liu2023lang2ltl, rabiei2025ltlcodegen,luo2025nl2spatial} to compile natural language commands into Linear Temporal Logic~(LTL) formulas using a large language model~(LLM). For instance, the household task above can be expressed as $\varphi = \ltlevent((\textit{chair} \lor \textit{couch}) \land \ltlevent\, \textit{tv})$, where $\ltlevent$ denotes the \textit{eventually} operator. The LTL formula is then converted into a Deterministic Finite
Automaton~(DFA)~\cite{de2013linear}, whose graph structure
enables principled sub-task decomposition. The DFA selects the
current navigation target through dynamic edge weights that
decrease when a target object is detected, enabling the system
to opportunistically reorder sub-tasks in real time to reduce the
overall travel distance. To support this, the system maintains
multi-channel value maps, one per target category, that
accumulate spatial knowledge in parallel so that no exploration
effort is lost when the planner switches targets.
At the \textit{navigation level}, we introduce directional
score. Rather than computing a single direction-agnostic score
from the full RGB observation, we
visually label frontier directions on the observation image and
query the VLM to score each direction independently. These
per-direction scores are temporally fused using scene-weighted
averaging and spatially smoothed through angle-based
interpolation to produce dense value maps.

Our contributions are as follows.
\begin{enumerate}
\item We propose ViTL, a zero-shot framework that enables robots to execute long-horizon natural language navigation tasks with complex temporal constraints.
\item We introduce a novel directional score method that leverages vision-language models to generate spatially-varying navigation value maps, improving single-target zero-shot navigation accuracy.
\item We conduct comprehensive experiments to evaluate ViTL's
capability for complex natural language navigation tasks.
\end{enumerate}

\section{RELATED WORK}

ObjectNav requires a robot to find an instance of a target object category in an unseen environment. Learned approaches train task-specific policies via reinforcement learning~\cite{chaplot2020semexp} or semantic map prediction~\cite{ramakrishnan2022poni}, but are restricted to closed-set categories and simulated data. Zero-shot methods avoid task-specific training by leveraging foundation models. CoW~\cite{gadre2022clip} explores the nearest frontier until an open-vocabulary detector finds the target. ESC~\cite{zhou2023esc} and L3MVN~\cite{yu2023l3mvn} use large language models to evaluate frontiers based on nearby object detections converted to text. VLFM~\cite{yokoyama2024vlfm} instead queries a vision-language model directly on the RGB observation to produce a cosine-similarity score, which is assigned uniformly across the field of view to build a value map. More recent methods enrich spatial reasoning through learned environment predictions~\cite{nie2025wmnav}, geometric cues~\cite{huang2024gamap}, and scene-graph representations~\cite{yin2024sg}.
All of these methods are designed for single-target navigation and cannot handle complex temporal dependencies. 

Executing multi-step tasks from natural language requires both high-level planning and low-level navigation. InstructNav~\cite{long2025instructnav} re-queries an LLM at every decision step for complex navigation tasks. 
However, it does not model temporal constraints between sub-tasks and provides no formal guarantees on task completion. NavGPT-2~\cite{zhou2024navgpt} aligns vision-language model features with a graph-based navigation policy for navigation tasks, but requires task-specific fine-tuning and does not operate in a zero-shot setting. SayPlan~\cite{rana2023sayplan} grounds LLM plans in robot affordances or scene graphs for long-horizon manipulation, but assumes known environments and does not address the navigation problem in unexplored spaces. Our approach instead uses a single offline LLM call to compile the task into an LTL formula, which is then executed by a DFA that provides provable temporal constraint satisfaction and supports dynamic replanning upon new object detections.

Translating natural language commands into formal temporal logic specifications is an active research area~\cite{chen2023nl2tl, fuggitti2023nl2ltl,xu2025nl2hltl2plan}. Lang2LTL~\cite{liu2023lang2ltl} fine-tunes language models to convert grounded natural language into LTL formulas for robot task planning. LTLCodeGen~\cite{rabiei2025ltlcodegen} leverages code-generation LLMs to produce syntactically correct LTL from natural language. 


Once an LTL formula is obtained, recent works use temporal logic to guide robot execution. SELP~\cite{wu2025selp} constructs a B\"{u}chi automaton to constrain LLM plan generation through token masking in known environments where all landmark locations are given; the automaton acts as an offline verifier and the robot executes the resulting plan without replanning. NL2HLTL2PLAN~\cite{fuggitti2023nl2ltl} extends NL-to-LTL translation to hierarchical specifications for multi-robot allocation, but similarly assumes known workspaces. In~\cite{taheri2025motion}, the authors address LTL planning under semantic uncertainty using a product automaton with value iteration over discrete grids, updating beliefs as labels are revealed. While related in handling unknown environments, their method operates on small discrete state spaces with probabilistic label priors rather than continuous visual observations. In contrast, our work targets zero-shot navigation in completely unknown continuous environments with no prior map. The compiled automaton serves as an online reactive controller that dynamically selects navigation targets as the robot discovers objects through visual perception, combined with VLM-based scoring to guide frontier exploration.


\section{PROBLEM FORMULATION}
\label{sec:problem}

Enabling robots to follow natural language commands to complete zero-shot long-horizon tasks in real-world environments remains challenging, as it requires extracting implicit temporal and logical constraints from natural language commands and executing multiple sub-tasks accordingly.

For example, a household service robot may receive a command such as: \textit{``Clean the chair or the couch, then turn on the tv.''} {Such a command defines a set of target object categories $\mathcal{C} = \{c_1, c_2, \ldots, c_n\}$ together with temporal constraints over them, such as \emph{sequencing} (chair or couch before tv) and \emph{disjunction} (chair or couch). We focus on the navigation component, assuming each sub-task is completed once the robot reaches the corresponding object, therefore, we reduce all action verbs (e.g., \emph{clean}, \emph{turn on}) to \emph{reach} objects.

{We consider a zero-shot setting where no prior map is available. At each timestep~$t$, the robot receives an egocentric observation $o_t = (I_t, D_t, \xi_t)$ consisting of an RGB image~$I_t$, a depth image~$D_t$, and the agent pose~$\xi_t$, and must output an action $a_t \in \{\texttt{move\_forward}, \texttt{turn\_left}, \texttt{turn\_right}, \texttt{stop}\}$ to progressively complete the task.}

{\textbf{Problem~1.} \textit{Given a natural-language command
specifying target object categories with temporal constraints, and without access to prior information of the environment, the robot must determine a sequence of actions based on observations~$o_t$ to reach all required targets while satisfying the temporal constraints specified by the command.}}

\begin{figure*}[t]
    \centering
    \includegraphics[width=0.9\textwidth]{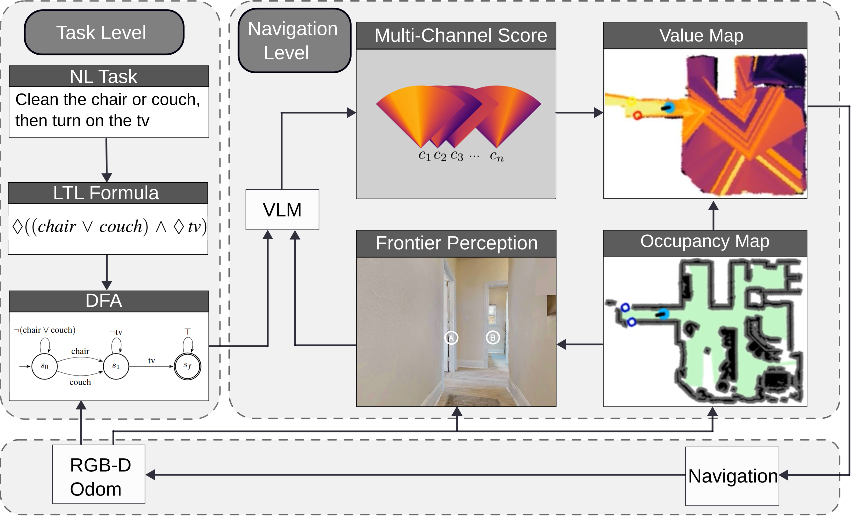}
    \caption{System overview of the ViTL framework. At the task level, a natural language command is compiled into an LTL formula via a single LLM call and then into a DFA that tracks progress toward task completion. When the object detector identifies a new target, the DFA updates the weights of corresponding edges and may trigger a target switch. At the navigation level, the VLM scores frontier directions in the current RGB observation for each target category, producing multi-channel value maps that are updated in parallel. The DFA selects the active target, and the agent navigates to the highest-score frontier in the corresponding value map channel. Depth and odometry observations maintain the occupancy map from which frontiers are extracted, and a point-goal policy converts selected waypoints into low-level actions.}
    \label{fig:system_structure}
\end{figure*}

\section{METHOD}
\label{sec:method}

We propose ViTL, a zero-shot navigation framework illustrated in Fig.~\ref{fig:system_structure}. Our approach is grounded in one observation: long-horizon commands require completing sub-tasks in an order that respects the embedded temporal constraints. In a zero-shot setting with no prior map, the system cannot fix this ordering in advance; it needs a task-level planner that incorporates new observations, selects the optimal sub-task online, and never violates the constraints while minimizing travel distance.

We observe that this problem maps naturally to graph
search over a DFA
compiled from LTL, where each path through the
graph corresponds to a valid execution ordering and
shortest-path search yields the optimal sub-task
(Sec.~\ref{sec:task_planner}). Each selected sub-task is
then executed by a navigation-level executor that performs
frontier-based exploration using directional score
and smooth value maps (Sec.~\ref{sec:nav_executor}).
When the task-level planner determines the current
sub-task, the navigation level executes it using the
corresponding value map channel.

\subsection{Task-Level Planner}
\label{sec:task_planner}


The task-level planner receives continuously updated environment
knowledge, the set of detected objects~$\mathcal{D}$, reached
objects~$\mathcal{R}$, and failed objects~$\mathcal{B}$, and
outputs the current optimal target~$c_k$. It operates in two
stages: compiling the natural language command into a formal
specification (Sec.~\ref{sec:nl_to_dfa}), and using graph search
over the resulting automaton to dynamically select and update the
active target as the environment state evolves
(Sec.~\ref{sec:target_selection}).

\subsubsection{From Natural Language to Automaton}
\label{sec:nl_to_dfa}


\textbf{Background: LTL and DFA.}
We use Linear Temporal Logic~\cite{de2013linear} to formally represent temporal constraints. An LTL formula over atomic propositions $\Pi$ is defined recursively as:
\begin{equation*}
\phi ::= \pi \mid \lnot \pi \mid \phi_1 \lor \phi_2 \mid \phi_1 \land \phi_2 \mid \ltlnext \phi \mid \phi_1 \ltluntil \phi_2 \mid \ltlevent \phi,
\end{equation*}
where $\pi \in \Pi$ is an atomic proposition, $\lnot$, $\land$, $\lor$ are Boolean operators, and $\ltlnext$ (next), $\ltluntil$ (until), $\ltlevent$ (eventually) are temporal operators. LTL formula can be compiled into a deterministic finite automaton (DFA) $\mathcal{A} = (S, \Sigma, \delta, s_0, F)$ using standard tools~\cite{duret2016spot}, where acceptance of a finite word certifies satisfaction of the formula. In our setting, each $\pi \in \Pi$ corresponds to a target object category, and DFA transitions are triggered when the robot reaches the associated object.

\textbf{Task compilation.}
To compile natural language commands 
into formal specifications, we prompt an LLM to extract the 
target categories $\mathcal{C}$ and temporal constraints, 
producing an LTL formula $\varphi$ over $\mathcal{C}$, following 
established NL-to-LTL translation 
approaches~\cite{liu2023lang2ltl}. This 
single offline LLM call replaces the repeated runtime queries 
required by LLM-based planners~\cite{long2025instructnav}, 
reducing both latency and the risk of cumulative reasoning 
errors. The resulting formula is then compiled into a 
deterministic finite automaton using standard 
tools~\cite{duret2016spot}, which serves as an online reactive 
controller throughout the mission. Consider the running 
example: after reducing all action verbs to \textit{reach}, 
the command becomes \textit{``Reach either a chair or a couch, then reach a tv.''} The LLM identifies $\mathcal{C} = \{\textit{chair}, \textit{couch}, \textit{tv}\}$ and produces:
\begin{equation*}
    \varphi = \ltlevent((\textit{chair} \lor \textit{couch}) \land \ltlevent\,\textit{tv})
    \label{eq:ltl_example}
\end{equation*}
The disjunction encodes a choice: the robot may reach either object to satisfy the first sub-task while the nested $\ltlevent$ enforces sequencing: the tv must be reached afterward. The LTL can be compiled to DFA 
(Fig.~\ref{fig:DFA}) which has two equally short paths from $s_0$: one via \textit{chair} and one via \textit{couch}, both leading to $s_1$, from which reaching \textit{tv} leads to the accepting state~$s_f$. This graph structure encodes all valid execution orderings, with branching representing choices and path length reflecting the number of remaining sub-tasks.

\begin{figure}[t]
    \centering
    \begin{tikzpicture}[>=Stealth, shorten >=1pt, auto, node distance=2.8cm,
        every state/.style={thick, minimum size=1cm}]
        \node[state, initial, initial text={}] (s0) {$s_0$};
        \node[state, right of=s0] (s1) {$s_1$};
        \node[state, accepting, right of=s1] (sf) {$s_f$};
        \path[->]
            (s0) edge[bend left=20] node {chair} (s1)
            (s0) edge[bend right=20] node[swap] {couch} (s1)
            (s1) edge node {tv} (sf)
            (s0) edge[loop above] node {$\neg(\text{chair} \lor \text{couch})$} (s0)
            (s1) edge[loop above] node {$\neg\text{tv}$} (s1)
            (sf) edge[loop above] node {$\top$} (sf);
    \end{tikzpicture}
    \caption{DFA compiled from $\varphi = \ltlevent((\textit{chair} \lor \textit{couch}) \land \ltlevent\,\textit{tv})$. State $s_0$ is initial; $s_f$ is accepting. Transitions are triggered by reaching the labeled target.}
    \label{fig:DFA}
\end{figure}
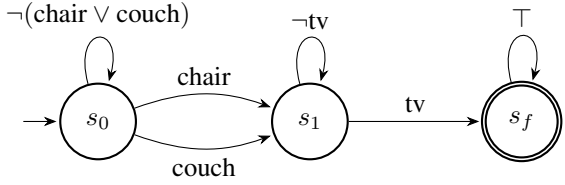

\subsubsection{Dynamic Target Selection and Replanning}
\label{sec:target_selection}


The compiled DFA $\mathcal{A}$ drives target selection throughout the mission. The planner maintains three sets that summarize the robot's evolving knowledge of the environment: detected objects~$\mathcal{D}$ (observed but not yet reached), reached objects~$\mathcal{R}$, and failed objects~$\mathcal{B}$ (targets the navigator failed to reach). The DFA state~$s$ advances only when the robot reaches a target ($s \leftarrow \delta(s, c_k)$). In case of update $\mathcal{B}$ or $\mathcal{R}$, the planner recomputes the shortest weighted path over the DFA to select the next target. Each edge $e$ in the DFA corresponds to reaching a specific 
target object $c_e$, and is assigned a weight $w(e)$ based on 
the robot's current knowledge:

\begin{equation}
    w(e) =
    \begin{cases}
        w_{\text{det}} & \text{if } c_e \in \mathcal{D} \\
        w_{\text{fail}} & \text{if } c_e \in \mathcal{B} \\
        1.0 & \text{otherwise}
    \end{cases}
    \label{eq:edge_weight}
\end{equation}
where $w_{\text{det}} \ll 1$ reflects that navigating to a detected target at a known location requires far less effort than exploring for an undetected one, and $w_{\text{fail}} \gg 1$ discourages revisiting targets that the navigator previously failed to reach.
The planner then finds the shortest weighted path $\rho^*$ through the graph search from the current state $s_{\text{curr}}$ to the accepting state $s_f$:
\begin{equation}
    \rho^* = \argmin_{\rho:\, s_{\text{curr}} \rightsquigarrow s_f} \sum_{e \in \rho} w(e)
    \label{eq:dijkstra}
\end{equation}
and the next navigation target $c_k$ is the label on the first edge of $\rho^*$. Infeasibility is detected in two cases. First, if the compiled DFA has no path from $s_0$ to an accepting state, the command is contradictory and is rejected before execution. Second, during execution, if $\sum_{e \in \rho^*} w(e) \geq w_{\text{fail}}$, every remaining path must traverse a failed target, and the task is declared infeasible. Note that constraints are enforced over \emph{reached} targets only; passing through a region without stopping does not trigger a transition, so merely traversing an area en route does not violate the specification.
Replanning is triggered whenever $\mathcal{D}$, $\mathcal{R}$, or $\mathcal{B}$ changes: edge weights are recomputed and the shortest path is updated. For instance, if the robot is exploring for \textit{chair} (not yet detected) and the detector spots a \textit{couch}, then \textit{couch} enters $\mathcal{D}$ while \textit{chair} has not: the edge $s_0 \xrightarrow{\textit{couch}} s_1$ receives weight~$w_{\text{det}}$ while the \textit{chair} edge retains weight~$1.0$, making the \textit{couch} path cheaper and triggering an immediate target switch.

\textbf{Multi-channel value maps.}
To support 
multi-target execution, 
the system maintains a
separate value map channel for each target category
from the natural language command:
\begin{equation}
    \mathcal{M} = \{\mathbf{V}^{(j)}\}_{j=1}^{n}
\end{equation}
where $\mathbf{V}^{(j)}$ is the value map for target $c_j$,
recording per-pixel navigation scores. All channels are updated
in parallel at every timestep using the procedure described in
Sec.~\ref{sec:nav_executor}. When the planner selects
target~$c_k$, it activates the corresponding
channel~$\mathbf{V}^{(k)}$ for the navigation-level executor,
which uses it to rank frontiers and guide exploration. 
When the planner
changes targets, the new channel already encodes spatial knowledge from prior exploration, reducing redundant re-exploration.

Algorithm~\ref{alg:DFA_nav} summarizes the procedure. The main loop alternates between detection-triggered replanning (lines~4--10), value map updates and agent navigation to the target (lines~11--20). If the current target has been detected, the agent navigates directly to its stored location $\ell(c_k)$; otherwise, it explores via the active value map channel. Upon reaching a target, the DFA state advances and a new target is selected; upon failure, the target is added to the failed set, edge weights are updated, and the planner replans.

\begin{algorithm}[t]
\caption{LTL-Based Navigation with Dynamic Replanning}
\label{alg:DFA_nav}
\begin{algorithmic}[1]
\STATE \textbf{Input:} LTL formula $\varphi$, target categories $\mathcal{C}$
\STATE $\mathcal{A} \leftarrow \textsc{CompileDFA}(\varphi)$; \; $s \leftarrow s_0$; \; $\mathcal{D}, \mathcal{R}, \mathcal{B} \leftarrow \emptyset$; \; $\mathcal{M} \leftarrow \{\mathbf{V}^{(j)} = \mathbf{0}\}_{j=1}^{n}$
\STATE $c_k \leftarrow \textsc{SelectTarget}(\mathcal{A}, s, \mathcal{D})$
\WHILE{$s \notin F$ \textbf{and} $t < T_{\max}$}
    \STATE $c_j \leftarrow \textsc{Detect}(I_t)$
    \IF{$c_j \notin \mathcal{D}$}
        \STATE $\mathcal{D} \leftarrow \mathcal{D} \cup c_j$; \; record location $\ell(c_j)$
        \STATE Update edge weights in $\mathcal{A}$~\eqref{eq:edge_weight}
        \STATE $c_k \leftarrow \textsc{SelectTarget}(\mathcal{A}, s, \mathcal{D})$ \hfill \textit{// replan}
    \ENDIF
    \STATE $\mathcal{M} \leftarrow \textsc{UpdateMaps}(\mathcal{M}, o_t, \mathcal{C})$ \hfill \textit{// Sec.~\ref{sec:nav_executor}}
    \IF{$c_k \in \mathcal{D}$}
        \STATE Navigate to $\ell(c_k)$
    \ELSE
        \STATE $f^* \leftarrow \text{Eq.}$~\eqref{eq:multi_frontier_selection} using $\mathbf{V}^{(k)}$
        \STATE Navigate to $f^*$ \hfill \textit{// explore}
    \ENDIF
    \IF{$\textsc{Reached}(c_k)$}
        \STATE $s \leftarrow \delta(s, c_k)$; \; $\mathcal{R} \leftarrow \mathcal{R} \cup \{c_k\}$
        \STATE $c_k \leftarrow \textsc{SelectTarget}(\mathcal{A}, s, \mathcal{D})$
\ELSIF{$\textsc{Failed}(c_k)$}
        \STATE $\mathcal{B} \leftarrow \mathcal{B} \cup \{c_k\}$; \STATE Update edge weights in $\mathcal{A}$ ~\eqref{eq:edge_weight}
        \STATE $c_k \leftarrow \textsc{SelectTarget}(\mathcal{A}, s, \mathcal{D})$ \hfill \textit{// replan}
    \ENDIF
\ENDWHILE
\RETURN $s \in F$
\end{algorithmic}
\end{algorithm}

\subsection{Navigation-Level Executor}
\label{sec:nav_executor}

Given a target $c_k$ and its corresponding value map channel
$\mathbf{V}^{(k)}$ selected by the task-level planner, the
navigation-level executor guides the robot toward it through
frontier-based exploration. We use a top-down occupancy map from depth observations and odometry, following~\cite{yokoyama2024vlfm}. Each boundary between explored and unexplored regions yields one frontier $f_i$, taken as its midpoint, a single 2D waypoint (Fig.~\ref{fig:system_structure}, Occupancy Map).

Existing value-map approaches compute a single direction-agnostic score from the full RGB observation, discarding spatial information about which region within the field of view is most relevant.
We instead introduce \textit{directional score}: the VLM
evaluates individual frontier directions, producing
spatially-varying value maps that more accurately capture which
regions are promising for reaching~$c_k$. Note that
single-target ObjectNav is the special case where
$\varphi = \ltlevent\, c_1$, reducing to one value
map channel with no task-level coordination. The following
subsections detail how directional scores are computed
(Sec.~\ref{sec:frontier_score}) and how the value maps are
constructed(Sec.~\ref{sec:smooth_map}).

\subsubsection{Directional Score}
\label{sec:frontier_score}

Frontiers extracted from the occupancy map are projected
into the current RGB frame based on the agent's pose. Each
projected frontier $f_i$ is assigned a letter label
$L_i \in \{A, B, C, \ldots\}$ and visually annotated on the
image (Fig.~\ref{fig:system_structure}, Frontier Perception).
An additional option $L_{N_f+1}$ represents ``explore elsewhere,''
serving as a calibration anchor that allows the VLM to distribute
probability mass away from visible frontiers when none is
promising.

We construct a multiple-choice prompt for the VLM:
\begin{quote}
\textit{``Among directions ($A$, $B$, $C$ ), or explore elsewhere ($D$), which should I choose to find a \{target\}? Answer with ONLY a single letter.''}
\end{quote}

The directional probability distribution is computed via softmax over the next-token logits for each option letter:
\begin{equation}
    P(f_i \mid I_t, c_k) = \frac{\exp(z_{L_i})}{\sum_{j=1}^{N_f+1} \exp(z_{L_j})}
    \label{eq:frontier_softmax}
\end{equation}
where $z_{L_i}$ is the logit of the token corresponding to letter $L_i$. This yields a probability for each frontier, providing both a ranking and a confidence measure.

We query the VLM for an overall scene relevance score:
\begin{quote}
\textit{``Does this view show paths or doorways that could lead to a \{target\}? Answer Yes or No.''}
\end{quote}
\begin{equation}
    S_{curr} = P(\text{Yes} \mid I_t, c_k) = \frac{\exp(z_{\text{Yes}})}{\exp(z_{\text{Yes}}) + \exp(z_{\text{No}})}
    \label{eq:scene_score}
\end{equation}
Each frontier's value blends the scene-level and direction-level signals:

\begin{equation}
    V_{\text{curr}}(f_i) =  S_{curr} \times P(f_i \mid I_t, c_k)
    \label{eq:v_curr}
\end{equation}

\begin{figure}[t]
    \centering
    \begin{subfigure}[b]{0.48\columnwidth}
        \centering
        \includegraphics[width=\textwidth]{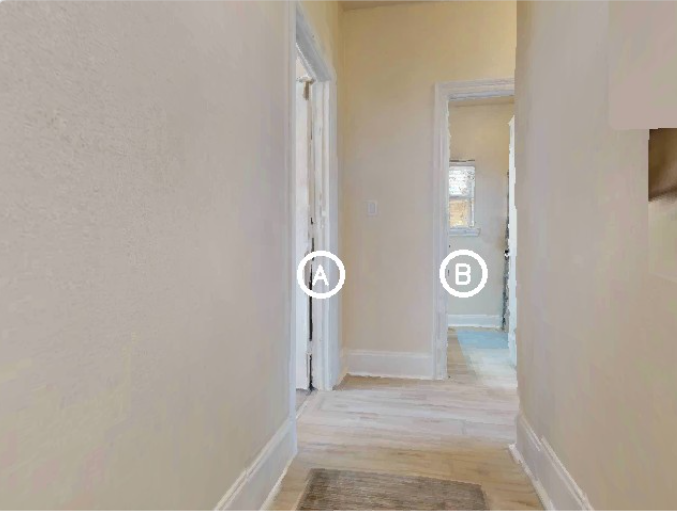}
        \caption{Observation with frontier directions labeled $A$, $B$.}
        \label{fig:value_map_obs}
    \end{subfigure}
    \hfill
    \begin{subfigure}[b]{0.48\columnwidth}
        \centering
        \includegraphics[width=\textwidth]{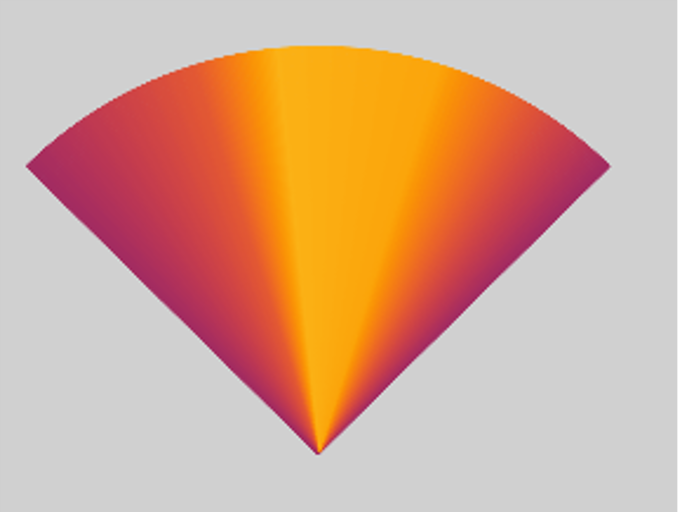}
        \caption{FOV cone projection and spatially smooth.}
        \label{fig:value_map_cone}
    \end{subfigure}
    \caption{Smooth value map generation. Discrete frontier scores (left) are converted into a smooth gradient within the FOV cone (right) via angle-based interpolation and temporally fused with prior observations.}
    \label{fig:value_map}
\end{figure}

\subsubsection{Smooth Value Map Construction}
\label{sec:smooth_map}

Given the per-frontier scores from Eq.~\eqref{eq:v_curr}, we
construct a 2D value map $\mathbf{V}$ that records a navigation
score at every explored pixel. At each timestep, frontier scores
are first temporally fused with historical observations at their
respective map locations, then spatially interpolated across the
Field-of-View (FOV) cone to yield a dense, smoothly varying gradient as shown in
Fig.~\ref{fig:value_map}.

\textbf{Temporal fusion.}
We maintain a scene score map $\mathbf{S}$, a 2D grid aligned with the value map, where each pixel stores the scene score $S_{curr}$ from when it was last observed. Each frontier $f_i$ occupies a specific pixel, at which we look up the stored value $\mathbf{V}_{\text{old}}(f_i)$ and scene score $\mathbf{S}_{\text{old}}(f_i)$. If the
pixel has never been observed
($\mathbf{S}_{\text{old}}(f_i) = 0$), the fused value is
the current score:
$\tilde{V}(f_i) = V_{\text{curr}}(f_i)$. Otherwise, the old and
new values are combined using scene-weighted averaging:
\begin{equation}
    \tilde{V}(f_i) = \frac{\mathbf{V}_{\text{old}}(f_i) +
    V_{\text{curr}}(f_i)}{\mathbf{S}_{\text{old}}(f_i) +
    S_{curr}} \cdot S_{curr}
    \label{eq:temporal_fusion}
\end{equation}
Because both terms are weighted by their respective scene scores,
observations from more informative viewpoints (higher $S_{curr}$)
naturally contribute more to the fused value.

\textbf{Spatial smoothing.}
The fused frontier values $\{\tilde{V}(f_i)\}$ are point
estimates at discrete locations. To produce a dense value map across the FOV, we compute each frontier's bearing angle $\theta_i$ relative to the camera's forward direction. Since each frontier $f_i$ is a single midpoint, $\theta_i$ is one scalar per frontier. We then insert \textit{virtual edge frontiers} at the FOV boundaries $\pm\theta_{\text{fov}}/2$ with baseline score $V_{\text{edge}} = \beta \cdot S_{curr}$, where $\beta$ is an edge attenuation factor. For each pixel $p$ inside the FOV cone
with bearing angle $\theta_p$, let $\theta_i'$ and
$\theta_{i+1}'$ be the two adjacent sorted angles such that
$\theta_i' \leq \theta_p \leq \theta_{i+1}'$. The smooth value
is obtained by linear interpolation:
\begin{equation}
    V_{\text{smooth}}(p) = v_i' + \frac{\theta_p - \theta_i'}
    {\theta_{i+1}' - \theta_i'} (v_{i+1}' - v_i')
    \label{eq:interpolation}
\end{equation}
where $\{\theta_i'\}$ and $\{v_i'\}$ are the angles and
fused scores in ascending angular order. The result is a smooth
gradient across the FOV cone
(Fig.~\ref{fig:value_map}b): high values concentrate near
promising frontiers and decay toward the edges. The smoothed
values are then written into $\mathbf{V}$ for pixels within
the current FOV, and the scene score map is updated accordingly.

\textbf{Frontier selection.}
The agent selects the next waypoint from the value map:
\begin{equation}
    f^* = \argmax_{f_i} \; \max_{p \in \mathcal{N}(f_i, r)}
    \mathbf{V}^{(k)}(p)
    \label{eq:multi_frontier_selection}
\end{equation}
where $\mathcal{N}(f_i, r)$ is the set of pixels within radius
$r$ of frontier $f_i$. 
The selected frontier serves as a waypoint for a pre-trained
depth-based PointNav policy, which
converts the relative goal position into low-level robot actions.

This cycle of map update, frontier selection and navigation repeats continuously: at each timestep the
occupancy map and all value map channels are updated with
new observations, frontiers are extracted, and the
highest-scoring frontier under $\mathbf{V}^{(k)}$ is
selected, and the agent navigates toward it, until the target object is detected, reached or failed.

\section{EXPERIMENTS}

\subsection{Experimental Setup}
We evaluate on the Habitat simulator~\cite{szot2021habitat} using the Habitat-Matterport 3D (HM3D) dataset~\cite{ramakrishnan2021hm3d}. The agent must navigate to an instance of a specified object category in an unseen environment using only egocentric RGB-D observations and pose estimates. We use the HM3D ObjectNav v2 validation split, which contains 2000 episodes across 20 scenes with six target categories: \{\textit{chair}, \textit{bed}, \textit{potted plant}, \textit{toilet}, \textit{tv monitor}, \textit{couch}\}. Each episode terminates after a maximum of $T_{\max} = 500$ steps.
The agent operates a discrete action space \{$\texttt{move\_forward}$ (0.25\,m), $\texttt{turn\_left}$ (30$^\circ$), $\texttt{turn\_right}$ (30$^\circ$), $\texttt{stop}$\}. 
An episode is considered successful if the agent calls \texttt{stop} within 1.0\,m of the target, and fail if it reaches the maximum steps.

\textbf{Evaluation Metrics.}
We report the two metrics: \textit{Success Rate} (SR), the success rate of navigation episodes, and \textit{Success weighted by Path Length} (SPL), measuring the success path efficiency if success SPL = $\frac{\text{optimal path}}{\text{actual path length}}$, otherwise SPL = 0. The higher value means a more efficient path.

\textbf{Implementation Details.}
We use LLaVA-NeXT-34B~\cite{liu2023visual} as the vision-language model, loaded with 8-bit quantization. For object detection, we employ YOLOv7-E6E~\cite{wang2023yolov7}. 
Low-level point-goal navigation is handled by a pre-trained depth-based PointNav policy~\cite{yokoyama2024vlfm}. The DFA is compiled from LTL specifications using the \texttt{lomap} library~\cite{ulusoy2013lomap}. 

\subsection{Navigation-Level Executor}
We first evaluate ViTL's capability in the navigation level, which is to solve the single-target ObjectNav problem.
Table~\ref{tab:single_target} compares ViTL's Navigation-Level Executor against other zero-shot object navigation methods on the HM3D ObjectNav v2 validation split (2000 episodes).
Our method achieves the success rate (54.1\%) and  SPL (30.9), demonstrating that directional score improves navigation reliability over other methods. ViTL achieves competitive results using LLaVA-NeXT-34B, an open-source model without fine-tuning, despite competing against methods that rely on significantly larger closed-source VLMs such as Gemini~1.5 Pro and GPT-4V. 

\begin{table}
\centering
\caption{Single-target ObjectNav results on HM3D v2 validation split (2000 episodes).}
\label{tab:single_target}
\begin{tabularx}{\columnwidth}{Xcc}
\toprule
Method & SR (\%) & SPL \\
\midrule
WMNav Gemini 1.5 Flash~\cite{nie2025wmnav} & 53.5 & 30.7 \\
WMNav Gemini 1.5 Pro~\cite{nie2025wmnav}   & 58.1 & 31.2 \\
GAMap~\cite{huang2024gamap}                 & 53.1 & 26.0 \\
SG-Nav-GPT~\cite{yin2024sg}                & 54.0 & 24.9 \\
VLFM~\cite{yokoyama2024vlfm}               & 52.4 & 30.8 \\
InstructNav (GPT4 + GPT4V)~\cite{long2025instructnav} & 56.0 & 22.5 \\
InstructNav (GPT4 + LLaVA 34B)~\cite{long2025instructnav} & 50.0 & 19.4 \\
\midrule
ViTL (LLaVA-34B) (ours)                    & 54.1 & 30.9 \\
\bottomrule
\end{tabularx}
\end{table}

\begin{figure}
    \centering

        \begin{subfigure}[b]{0.48\columnwidth}
        \centering
        \includegraphics[width=\textwidth]{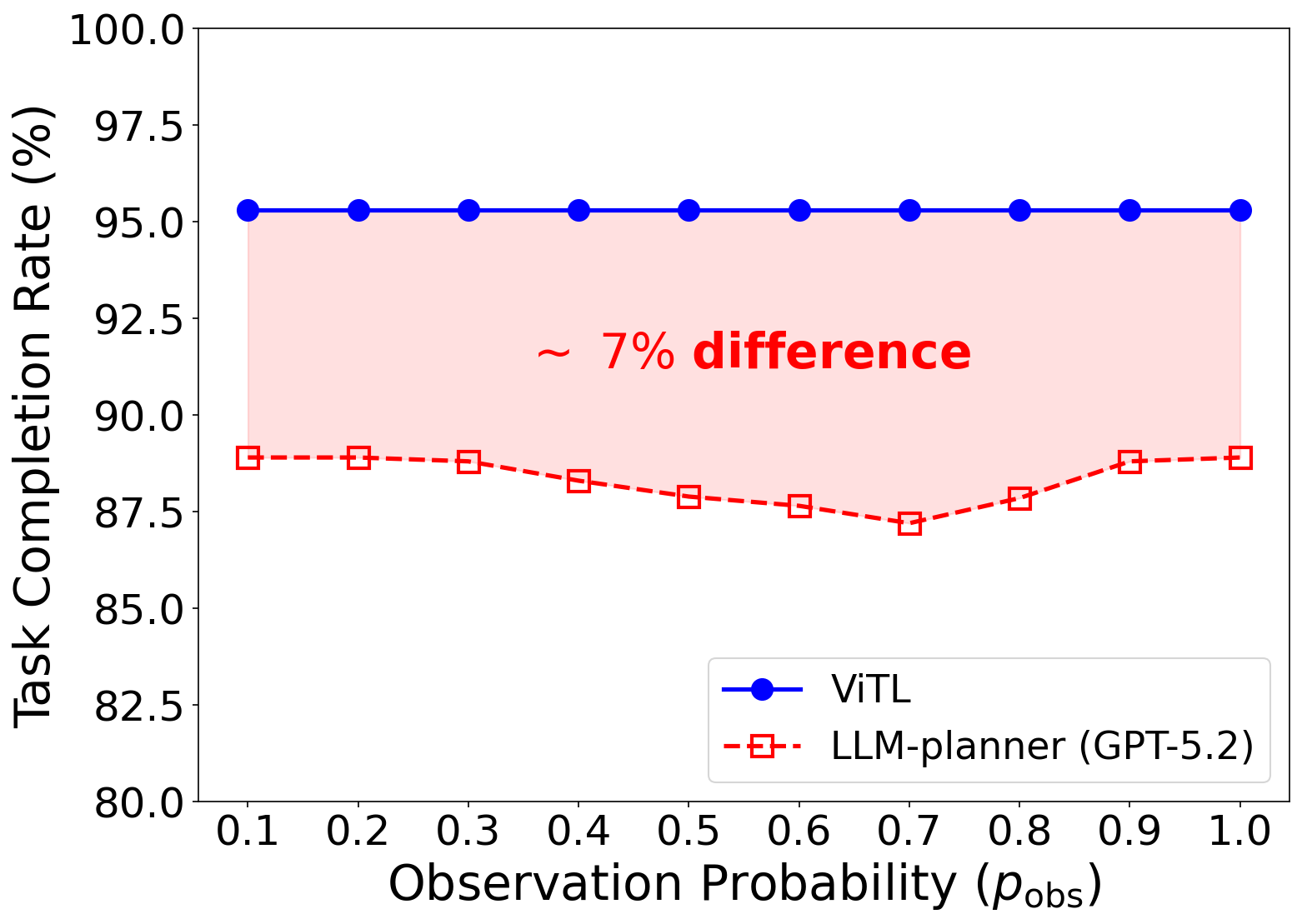}
        \caption{Task completion rate.}
        \label{fig:planning_success}
    \end{subfigure}
    \hfill
    \begin{subfigure}[b]{0.48\columnwidth}
        \centering
        \includegraphics[width=\textwidth]{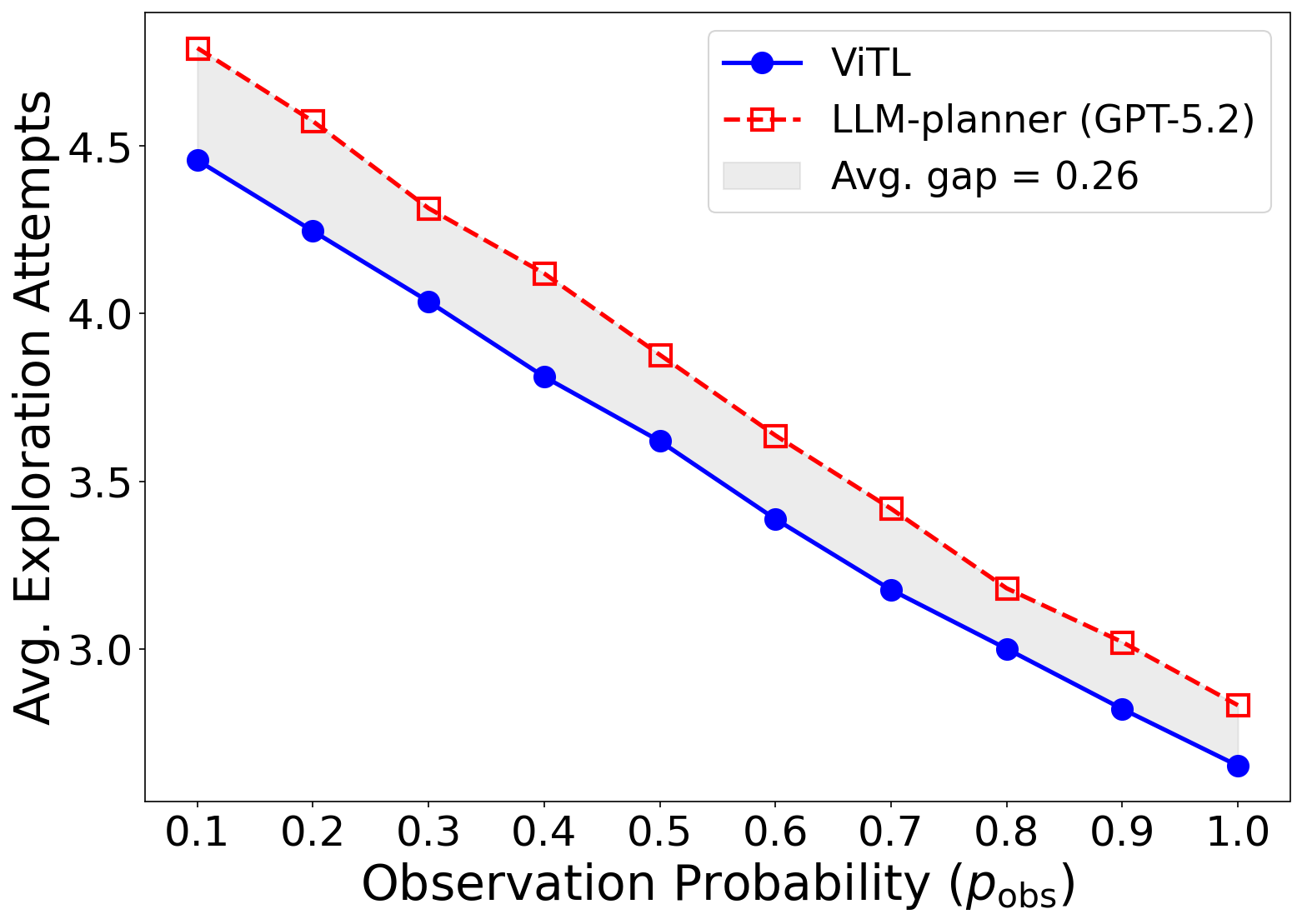}
        \caption{Avg. exploration attempts.}
        \label{fig:planning_attempts}
    \end{subfigure}
    \caption{Task-level planning performance across observation probabilities $p_{\text{obs}}$ on 2000 tasks. ViTL achieves higher task completion rate with fewer exploration attempts.}
    \label{fig:planning_sweep}
\end{figure}

\subsection{Task-Level Planner}
\label{sec:exp_planner}

We next evaluate ViTL's task-level planner: given a natural
language command and a stream of environment updates
($\mathcal{D}$, $\mathcal{R}$, $\mathcal{B}$), can the planner
reliably select targets that satisfy temporal constraints while
minimizing total travel cost? Since no existing zero-shot
navigation system supports temporally constrained multi-target
tasks, we compare against an LLM-based baseline
(\textit{LLM-planner}) on 2000 generated natural language
navigation tasks.

\textbf{Benchmark and evaluation protocol.}
We build a benchmark of 2000 natural language navigation tasks,
each paired with a ground-truth LTL formula. Each formula is
constructed from a lifted template: an abstract temporal
structure with placeholder propositions that is grounded by
sampling concrete objects from~$\mathcal{C}$. Templates range up to complex formulas
with disjunctive branches and \textit{until} constraints
(up to 6 targets). For example, \textit{``Reach a bed, then a
tv, then a couch; or reach a toilet, then a chair''} yields
$\ltlevent(\textit{bed} \land \ltlevent(\textit{tv} \land
\ltlevent\,\textit{couch})) \lor \ltlevent(\textit{toilet}
\land \ltlevent\,\textit{chair})$.

To enable efficient large-scale evaluation independent of low level navigation, we model the single-target executor
probabilistically. If the planner selects a previously observed
target, the robot reaches it directly. Otherwise, it explores:
with probability $p_{\text{obs}}$ it finds the intended target,
and with probability $1 - p_{\text{obs}}$ it encounters a
different object, triggering replanning. Each
exploration attempt corresponds to a full navigation episode, so
the total number of attempts reflects overall travel cost in real world. We
sweep $p_{\text{obs}} \in \{0.1, 0.2, \ldots, 1.0\}$ to assess the
planner: low values simulate cluttered environments
where exploration frequently reveals unintended objects, while
high values model simpler scenes.

\textbf{Baseline.}
Since the task is specified in natural language, the most
intuitive baseline is to use an LLM directly as a task-level
planner. The \textit{LLM-planner} does exactly this: at every
decision point, it sends a prompt containing the natural language
command and the current environment state
($\mathcal{D}$,~$\mathcal{R}$,~$\mathcal{B}$) to the LLM, which
returns the next target to pursue. The LLM is re-queried whenever
the environment state changes, and the process repeats until it
outputs ``Done'', indicating that the task is complete or infeasible. The prompt template is shown below. 

ViTL also uses an LLM, but queries it only once to translate the natural language command into an LTL formula, which is then compiled into a DFA and used for graph-based target selection. To ensure a fair comparison, both methods share the same LLM (GPT-5.2), environment information, and navigation-level executor, so any difference in performance is attributable solely to the planner.

\begin{figure}[t!]
\begin{tcolorbox}[
    colback=gray!5,
    colframe=gray!60,
    fonttitle=\bfseries\small,
    title=LLM-Planner Prompt,
    boxrule=0.5pt,
    arc=2pt,
    left=4pt, right=4pt, top=2pt, bottom=2pt
]
\ttfamily\small
\textbf{Role:} You are a navigation planner. Determine the next target object for an agent to satisfy the task as fast as possible.\\ [3pt]
\textbf{Task:} \{Task description in natural language\}\\ [3pt]
\textbf{Rules:} Follow the temporal ordering implied by the task. Revisiting objects is allowed. Output exactly: \texttt{TARGET: <object>} or \texttt{"DONE"}.\\ [3pt]
\textbf{Current State:}\\
\hspace*{1em}$\bullet$ Observed (known location, not yet reached): \{...\}\\
\hspace*{1em}$\bullet$ Reached (in order): [...]\\
\hspace*{1em}$\bullet$ Failed (unreachable): \{...\}
\end{tcolorbox}
\end{figure}

\textbf{Results.}
Fig.~\ref{fig:planning_sweep} shows the results across
$p_{\text{obs}} \in \{0.1, 0.2, \ldots, 1.0\}$. ViTL reduces
the problem of multi-step temporal reasoning to NL-to-LTL
translation, which achieves 95.3\% accuracy (1906/2000 correct
formulas). Given a correct translation, the task-level planner in ViTL
\textit{guarantees} 100\% task completion across all
$p_{\text{obs}}$ values, as task execution reduces to graph
search. 
The figure reports end-to-end success rate, counting incorrect NL-to-LTL translations as failures; thus ViTL's success rate is capped at the 95.3\% translation accuracy.
We note that fine-tuned
compact models such as T5-base also achieve near-perfect
translation accuracy~\cite{liu2023lang2ltl}, indicating that this
step is not a bottleneck and can be performed reliably with modest
computational resources.

In contrast, \textit{LLM-planner} must perform temporal reasoning
incrementally at every decision point, and achieves only 88.3\%
completion with GPT-5.2. Its ${\sim}11\%$ failure rate
remains roughly constant across $p_{\text{obs}}$, confirming
that failures stem from reasoning errors, temporal constraint
violations or premature termination. ViTL also consistently requires fewer exploration
attempts (Fig.~\ref{fig:planning_attempts}). We verify that
this planning-level advantage translates into real navigation
savings in the full system evaluation later.


\textbf{Case studies.}
We present two examples from the benchmark that illustrate
typical \textit{LLM-planner} failures.

\textit{Case 1: Temporal constraint violation.}

\noindent\fbox{\parbox{0.95\columnwidth}{\small
\textbf{Task:} ``Reach a toilet and then a bed; or reach a couch
and then a chair; or reach a potted plant and then a tv. Do not
reach a tv until a bed is reached, and do not reach a toilet
until a couch is reached, and do not reach a couch until a tv is
reached.''

\textbf{Translated LTL:}
$\varphi = \ltlevent(\textit{toilet} \land
\ltlevent\,\textit{bed}) \lor
\ltlevent(\textit{couch} \land \ltlevent\,\textit{chair}) \lor
\ltlevent(\textit{plant} \land \ltlevent\,\textit{tv})$
$\land\; (\neg\textit{tv}\;\mathcal{U}\;\textit{bed})
\;\land\; (\neg\textit{toilet}\;\mathcal{U}\;\textit{couch})
\;\land\; (\neg\textit{couch}\;\mathcal{U}\;\textit{tv})$
}}
\vspace{0.15cm}

\begin{table}[h]
\centering
\small
\caption{Case 1 Path}
\label{table:case1}
\begin{tabularx}{\columnwidth}{@{}c|>{\centering\arraybackslash}X>{\centering\arraybackslash}X@{}}
\toprule
Step & ViTL & LLM-planner \\
\midrule
1 & bed             & tv \\
2 & potted plant    & bed \\
3 & tv              & couch \\
4 & ---             & chair \\
\midrule
Task completion & Yes   & No \\
\bottomrule
\end{tabularx}
\end{table}

\noindent In Table.~\ref{table:case1}, ViTL identifies bed $\rightarrow$ potted plant
$\rightarrow$ tv as the shortest path, correctly
respecting all temporal constraints. The \textit{LLM-planner}
reaches tv at the first step, directly violating the constraint
$\neg\textit{tv}\;\mathcal{U}\;\textit{bed}$ (tv must not be
reached before bed). Despite reaching 4~targets, the task
remains unsatisfied because the temporal constraint was violated.

\textit{Case 2: Premature termination.}

\vspace{0.15cm}
\noindent\fbox{\parbox{0.95\columnwidth}{\small
\textbf{Task:} ``Reach a couch and then a toilet and then a bed
and then a chair; or reach a bed and then either a potted plant
or a toilet; or reach a tv and then a potted plant and then a
couch. Do not reach a couch until a toilet is reached, and do
not reach a bed until a tv is reached, and do not reach a potted
plant until a chair is reached.''

\textbf{Translated LTL:}
$\varphi = \ltlevent(\textit{couch} \land
\ltlevent(\textit{toilet} \land \ltlevent(\textit{bed} \land
\ltlevent\,\textit{chair}))) \lor
\ltlevent(\textit{bed} \land \ltlevent(\textit{plant} \lor
\textit{toilet})) \lor \ltlevent(\textit{tv} \land
\ltlevent(\textit{plant} \land \ltlevent\,\textit{couch}))$
$\land\; (\neg\textit{couch}\;\mathcal{U}\;\textit{toilet})
\;\land\; (\neg\textit{bed}\;\mathcal{U}\;\textit{tv})
\;\land\; (\neg\textit{plant}\;\mathcal{U}\;\textit{chair})$

}}

\begin{table}[h]
\centering
\small
\caption{Case 2 Path}
\label{table:case2}
\begin{tabularx}{\columnwidth}{@{}c|>{\centering\arraybackslash}X>{\centering\arraybackslash}X@{}}
\toprule
Step & ViTL & LLM-planner \\
\midrule
1 & tv              & tv \\
2 & bed             & toilet \\
3 & toilet          & couch \\
4 & ---             & bed \\
5 & ---             & chair \\
\midrule
Task completion & Yes   & No \\
\bottomrule
\end{tabularx}
\end{table}

\noindent In Table.~\ref{table:case2}, ViTL identifies that the path with
tv $\rightarrow$ bed $\rightarrow$ toilet is the shortest satisfying path and
terminates correctly after 3~targets. The \textit{LLM-planner}
instead pursues a longer branch, visiting 5~targets,
and falsely declares the task complete despite the specification remaining unsatisfied. 

Together, these cases illustrate two fundamental limitations of
\textit{LLM-planner}: temporal constraint violations during
execution, and premature declaration of task completion. ViTL
is immune to both, as the DFA enforces constraint satisfaction
and accepts only at accepting states.

\subsection{Full System Evaluation}

End-to-end success rate is not the most informative metric for evaluating the task-level planner.
With single-target ObjectNav succeeding at approximately 54\%, a sequential 3-target task would succeed
roughly 15\% of the time even with a perfect planner, dropping exponentially with more targets.
The dominant failure mode in this regime is navigation-level, not planning-level.
To isolate planning quality, we select two \mbox{HM3D} scenes
(\texttt{ziup5kvtCCR} and \texttt{svBbv1Pavdk}) where the single-target navigator achieves near-90\% success rate,
substantially reducing navigation-level noise.
We run episodes in these scenes and compare path lengths on 50 mutually completed episodes (cases
where both methods successfully reached all required targets) so that any difference is attributable to target selection quality.

Both methods use the same navigation-level executor.
On mutually completed episodes, ViTL achieves an average path length of \textbf{25.45}\,m compared to \textbf{30.05}\,m
for \textit{LLM-planner}, a \textbf{15.31}\% reduction.
This confirms that fewer exploration attempts translate into shorter navigation paths.
\section{CONCLUSION}

We presented ViTL, a zero-shot navigation framework that enables
robots to complete long-horizon natural language navigation tasks
with temporal constraints. At the task level, ViTL translates
natural language commands into LTL formulas and compiles them into
a DFA, whose graph structure guarantees temporal constraint
satisfaction and identifies the
optimal execution order, adapting in real time as environment
information changes. At the navigation level, directional score
produces spatially-varying value maps that improve single-target
navigation accuracy and efficiency. Experiments on HM3D
demonstrate competitive single-target performance and reliable
completion of natural language navigation tasks, where LLM-based planners can violate temporal
constraints or fail to identify optimal execution paths.
Our framework currently assumes that each sub-task reduces to
reaching a single object. Extending ViTL to handle richer
sub-task semantics such as manipulation or interaction is a
promising direction for future work.

\bibliographystyle{IEEEtran}
\bibliography{reference}

\end{document}